\documentclass[sigconf]{acmart}

\renewcommand\footnotetextcopyrightpermission[1]{} % removes footnote with conference information in first column
\setcopyright{none}

\usepackage{graphicx}
\usepackage{tabularx}
\usepackage{booktabs}
\usepackage{float}

\usepackage{color}
\usepackage{siunitx}
\usepackage{multirow}
\usepackage{xcolor}
\usepackage{amsmath}
\usepackage{bbm}
\usepackage[utf8]{inputenc}
\usepackage[english]{babel}
\usepackage[T1]{fontenc}
\usepackage[autostyle=true]{csquotes}
\usepackage{caption}
\usepackage{subcaption}
\usepackage{balance}
\usepackage[nolist,nohyperlinks]{acronym}

\DeclareMathOperator*{\argmin}{arg\,min}

\newcolumntype{Y}{>{\raggedleft\arraybackslash}X}
\newcolumntype{R}{>{\hsize=0.5\hsize}Y}

\AtBeginDocument{%
  \providecommand\BibTeX{{%
    \normalfont B\kern-0.5em{\scshape i\kern-0.25em b}\kern-0.8em\TeX}}}

\acmConference{}
\acmBooktitle{}
\acmSubmissionID{2849}

\begin{document}

\begin{acronym}
\acro{EMA}{Exponential Moving Average}
\acro{CNN}{Convolutional Neural Network}
\acro{MLP}{Multilayer Perceptron}
\acro{GIoU}{Generalized Intersection over Union}
\acro{COCO}{Common Objects in Context}
\acro{PoPArt}{Poses of People in Art}
\acro{AP}{Average Precision}
\acro{AR}{Average Recall}
\acro{DETR}{Detection Transformer}
\acro{NDCG}{Normalized Discounted Cumulative Gain}
\acro{PVT}{Pyramid Vision Transformer}
%\acro{YOLO}{You Only Look Once}
\acro{HRNet}{High-Resolution Net}
\end{acronym}

%%%%%%%%% TITLE
% \title{Semi-supervised Pose Estimation in Artworks}
% \title{Semi Supervised Pose Estimation in Images for Art Retrieval}
% \title{Finding Image Adaptations with Pose Estimation in Artworks}
% \title{Semi-supervised Human Pose Estimation in Art-historical Imagery}
\title[Semi-supervised Human Pose Estimation in Art-historical Images]{Semi-supervised Human Pose Estimation in Art-historical Images}

\author{Matthias Springstein}
\email{matthias.springstein@tib.eu}
\affiliation{%
  \institution{\footnotesize TIB -- Leibniz Information Centre for Science and Technology}
  \city{Hannover}
  \country{Germany}
}

\author{Stefanie Schneider}
\email{stefanie.schneider@itg.uni-muenchen.de}
\affiliation{%
  \institution{\footnotesize Ludwig Maximilian University of Munich}
  \city{Munich}
  \country{Germany}
}

\author{Christian Althaus}
\email{christian.althaus@gmx.eu}
\affiliation{%
  \institution{\footnotesize TIB -- Leibniz Information Centre for Science and Technology}
  \city{Hannover}
  \country{Germany}
}

\author{Ralph Ewerth}
\email{ralph.ewerth@tib.eu}
\affiliation{%
  \institution{\footnotesize TIB -- Leibniz Information Centre for Science and Technology\\L3S Research Center, Leibniz University}
  \city{Hannover}
  \country{Germany}
}

\renewcommand{\shortauthors}{Springstein et al.}

\begin{abstract}
Gesture as \enquote*{language} of non-verbal communication has been theoretically established since the 17th century. 
However, its relevance for the visual arts has been expressed only sporadically. 
This may be primarily due to the sheer overwhelming amount of data that traditionally had to be processed by hand. 
With the steady progress of digitization, though, a growing number of historical artifacts have been indexed and made available to the public, creating a need for automatic retrieval of art-historical motifs with similar body constellations or poses. 
Since the domain of art differs significantly from existing real-world data sets for human pose estimation due to its style variance, this presents new challenges. 
In this paper, we propose a novel approach to estimate human poses in art-historical images. 
In contrast to previous work that attempts to bridge the domain gap with pre-trained models or through style transfer, we suggest semi-supervised learning for both object and keypoint detection. 
Furthermore, we introduce a novel domain-specific art data set that includes both bounding box and keypoint annotations of human figures. 
Our approach achieves significantly better results than methods that use pre-trained models or style transfer.
\end{abstract}
\maketitle


\section{Introduction}
\label{sec:intro}

\begin{figure*}
\centering
\begin{subfigure}{.247\textwidth}
  \centering
  \includegraphics[width=\linewidth]{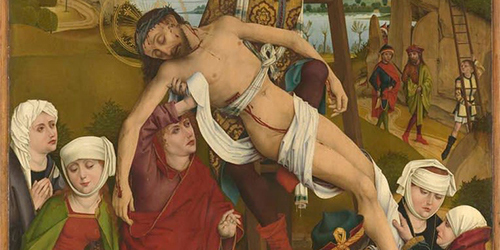}
  \caption{}
\end{subfigure}%
\hfill
\begin{subfigure}{.247\textwidth}
  \centering
  \includegraphics[width=\linewidth]{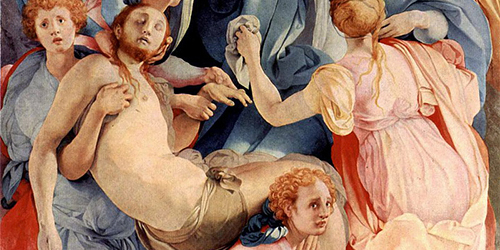}
  \caption{}
  \label{fig:deposition-pontormo}
\end{subfigure}%
\hfill
\begin{subfigure}{.247\textwidth}
  \centering
  \includegraphics[width=\linewidth]{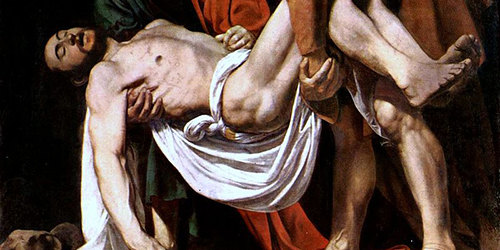}
  \caption{}
\end{subfigure}%
\hfill
\begin{subfigure}{.247\textwidth}
  \centering
  \includegraphics[width=\linewidth]{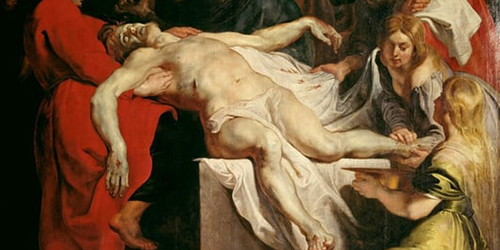}
  \caption{}
\end{subfigure}%
%\vspace{-9pt}
\caption{Depictions of Christ's deposition from the cross  with slightly varying poses by (a) Hans Pleydenwurff, 1465; (b) Pontormo, 1525--1528; (c) Caravaggio, 1603--1604; and (d) Peter Paul Rubens, ca. 1612. All images are in the public domain.
\vspace{4pt}
%Four depictions of Christ's deposition from the cross with slightly varying poses. (a) Hans Pleydenwurff, \textit{Crucifixion of the Hof Altarpiece}, 1465; (b) Jacopo da Pontormo, \textit{The Deposition from the Cross}, 1525--1528; (c) Caravaggio, \textit{The Entombment of Christ}, 1603--1604; (d) Peter Paul Rubens, \textit{The Entombment}, c. 1612.
}
\label{fig:deposition}
\end{figure*}

As \enquote*{language} of non-verbal communication, gesture %and posture have been
has been theoretically established since the 17th century \cite{Knowlson1965}. 
Its relevance for the visual arts, however, has so far been expressed at most sporadically \cite{Barasch1987}: e.g., symbolically-performatively on the basis of the medieval law-book manuscript of the Heidelberg \textit{Sachsenspiegel} \cite{vonAmira1905}, as the antiquity-receiving \enquote*{Pathosformel} \cite{Warburg1905, Warburg1914}, or as a status identifier exemplified in Roman sculpture \cite{Brilliant1963}. 
This selectivity may be primarily due to the sheer overwhelming amount of data that traditionally had to be processed manually. 
Driven by the steady progress of digitization, though, an increasing quantity of historical artifacts has been indexed and made freely available to the public online in recent decades. 
As a result, art historians can draw on ever larger collections of art-historical imagery to demonstrate the formulaic recapitulation of motifs with significant gesture or pose;\footnote{For simplicity, we hereinafter do not distinguish between the terms \enquote*{gesture,} \enquote*{posture,} and \enquote*{pose.} Instead, we use the term \enquote*{pose} for any kind of physical expression.} as exemplified by Christ's deposition from the cross in Figure~\ref{fig:deposition}. 
This is accompanied by a need for search engines that retrieve human figures with similar poses, facilitating the search for objects relevant to the individual scholar. 
It would thus become feasible to examine dominant pose types or time-dependent bodily phenomena on a large scale, as they were characteristic in Mannerism through the overlengthening of limbs, e.g., in Jacopo da Pontormo's work (Figure~\ref{fig:deposition-pontormo}). 
Intra- as well as inter-iconographic recurrent motifs, whose radically altered semantics are disconcerting, might be thoroughly discussed in this context. 
%In the following, we use the terms \enquote*{gesture} and \enquote*{posture} synonymously in art-historical contexts, but prefer the term \enquote*{pose} in computer science contexts. %related content.
To date, however, only few approaches exist for human pose estimation in art-historical images, possibly due to the lack of a sufficiently large domain-specific data set. To deal with this issue, one type of approaches uses pre-trained models, but without adapting them to the new domain~\cite{DBLP:conf/eccv/MadhuMKBMC20, DBLP:conf/icdar/JenicekC19}, while others apply style transfer to real-world data sets to obtain domain-specific training data~\cite{DBLP:journals/corr/abs-2012-05616}, or fine-tune pre-trained models using small, keypoint-level annotated data sets ~\cite{DBLP:journals/corr/abs-2012-05616}.

In this paper, we propose a novel approach to quantitatively systematize the exploration of pose types in visual art utilizing semi-supervised learning. % that utilizes state-of-the-art semi-supervised learning techniques. 
We suggest a two-stage approach based on two Transformer models: the first model detects bounding boxes of human figures and the second model predicts the keypoints of each box. 
We adapt a semi-supervised learning technique to reduce the performance loss caused by the shift between existing real-world data sets and the art domain, and to reduce the need to explicitly annotate a large amount of art-historical data. 
Our main contributions are as follows: 
(1) for object and keypoint detection, we suggest to combine semi-supervised pipelines through a two-step approach built on Transformer models with a teacher-student design; 
(2) to properly test our approach, we introduce a sufficiently large art-historical data set with both bounding box and keypoint annotations of human figures in $22$ depiction styles; 
(3) in contrast to previous work, we show that %the synthesizing real-world imagery 
the synthetic generation of seemingly \enquote*{realistic} art imagery inadequately reflects the stylistic diversity of historical artifacts. 
For both detection steps, the incorporation of manually labeled domain-specific material is performance-wise still required in the training and test phases.
The source code and models are publicly available.\footnote{\url{https://github.com/TIBHannover/iart-semi-pose}, last accessed on \today.}

The rest of the paper is structured as follows. Section~\ref{sec:rw} reviews related work on pose estimation and semi-supervised learning. In Section~\ref{chp:method}, we describe our pose estimator and its extension to a semi-supervised approach. 
% In Section~\ref{chp:exp}, we introduce a new data set and report experiments with various data sets.
In Section~\ref{chp:exp}, we introduce our data sets and report on the ablation studies performed.
Section~\ref{chp:study} presents a user study to evaluate retrieval results from a human perspective. We conclude with Section~\ref{chp:conc} and outline areas of future work.

\section{Related Work}
\label{sec:rw}

As with many other computer vision tasks, there has been steady progress in human pose estimation over recent years, particularly with the continued development of increasingly advanced deep learning models and self-supervised learning techniques.

\textbf{Human pose estimation} deals with the localization of a person's skeleton by detecting associated keypoints, i.e., \textit{skeleton coordinates} that mostly correspond to joint points of elbows, shoulders, etc.~\cite{DBLP:conf/cvpr/0012WZXXT21, DBLP:conf/cvpr/0009XLW19, DBLP:conf/cvpr/KreissBA19,  DBLP:conf/cvpr/ChengXWSH020, DBLP:journals/pami/CaoHSWS21, DBLP:conf/eccv/XiaoWW18, DBLP:conf/cvpr/ZhangZD0Z20}.
%In \textbf{human pose estimation}, persons are detected in an input image and their associated keypoints are determined, i.e., \textit{skeleton coordinates} that mostly correspond to joint points, e.g., elbows, knees, or shoulders~\cite{DBLP:conf/cvpr/0012WZXXT21, DBLP:conf/cvpr/0009XLW19, DBLP:conf/cvpr/KreissBA19,  DBLP:conf/cvpr/ChengXWSH020, DBLP:journals/pami/CaoHSWS21, DBLP:conf/eccv/XiaoWW18, DBLP:conf/cvpr/ZhangZD0Z20}. 
The problem can be solved in two ways. The \textit{top-down} approach first detects persons, indexes them with bounding boxes, and then determines keypoints for each person~\cite{DBLP:conf/cvpr/0012WZXXT21, DBLP:conf/cvpr/0009XLW19, DBLP:conf/cvpr/KreissBA19}; while the \textit{bottom-up} approach first detects keypoints, and then merges them to simultaneously identify persons and their basic pose~\cite{DBLP:journals/pami/CaoHSWS21, DBLP:conf/cvpr/ChengXWSH020, DBLP:conf/eccv/PapandreouZCGTM18}. 
Current work on the respective strategies shows that top-down methods generally yield better results, but at the cost of computational complexity~\cite{DBLP:conf/cvpr/ChengXWSH020}. 
Two-stage estimation makes the runtime linearly dependent on the number of detected persons in a scene, as the individual instances are cropped, and thus more forward steps are required for keypoint recognition. 
However, since there is no real-time requirement for the domain considered here, runtime is of secondary importance. Further differences result from the prediction of the individual keypoints.
Heatmap-based methods generate a dense likelihood map for the individual joints of the pose~\cite{DBLP:conf/cvpr/0009XLW19}, whereas regression models directly predict coordinates of the individual components and optimize them~\cite{DBLP:conf/cvpr/0012WZXXT21}. 
While heatmap-based methods tend to perform better, the advantage of regression-based models is that they require fewer pre- and post-processing steps~\cite{DBLP:conf/cvpr/0012WZXXT21}. 
As these models can be more easily used in our proposed semi-supervised technique, we also take advantage of this.

Few studies specifically address the estimation of human poses in art-historical images. This may be due to the fact that domain-specific data sets are usually only superficially indexed~\cite{painter, artigo, DBLP:conf/bmvc/KarayevTHADHW14} and rarely include fine-grained annotations at the level of concrete image details~\cite{DBLP:conf/eccv/GarciaV18,DBLP:conf/mm/MaoCS17, DBLP:journals/corr/abs-1906-00901, DBLP:journals/tomccap/StrezoskiW18}. 
A publicly accessible data set that contains poses of human figures in artworks does not yet exist. 
Relevant previous work employs different approaches to deal with the lack of annotated training data: they (1) analyze only self-annotated data sets, without training models or performing inference~\cite{DBLP:conf/eccv/ImpettS16}; 
(2) use trained pose estimators from another domain without adaptation \cite{DBLP:conf/eccv/MadhuMKBMC20, DBLP:conf/icdar/JenicekC19}; 
(3) apply style transfer to real-world data sets to close the domain gap~\cite{DBLP:journals/corr/abs-2012-05616}; 
or (4) leverage small, keypoint-level annotated data sets to fine-tune pre-trained models~\cite{DBLP:journals/corr/abs-2012-05616}.

%\textbf{Semi-supervised learning} was born out of the necessity that the performance of current deep learning models depends significantly on the amount of available data. 
\textbf{Semi-supervised learning} aims to exploit a (potentially large) set of unlabeled data in addition to a (typically small) set of labeled data to improve the resulting model.
%Typically, only a small portion of the training data is annotated in semi-supervised methods.
To use the rest of the material during training, pseudo-labels are either generated~\cite{DBLP:conf/cvpr/XieLHL20, lee2013pseudo}, or integrated into the loss with consistency regularization~\cite{DBLP:conf/iclr/LaineA17, DBLP:journals/pami/MiyatoMKI19}. 
One type of state-of-the-art methods uses a teacher-student approach. 
During training, an image is fed into a teacher model, which then generates a label for a student model that is being trained. 
The teacher model update can be iteratively selected from a previously trained student model~\cite{DBLP:conf/cvpr/XieLHL20}, or the teacher is an \ac{EMA} of the student~\cite{DBLP:conf/iclr/TarvainenV17}. 
Another type of semi-supervised methods uses data augmentation to generate better feedback signals for unlabeled data, or combines pseudo-label generation and consistency regularization~\cite{DBLP:conf/nips/BerthelotCGPOR19, DBLP:conf/iclr/BerthelotCCKSZR20, DBLP:conf/nips/SohnBCZZRCKL20}. 
Similar to semi-supervised classification, semi-supervised localization provides support for consistency regularization~\cite{DBLP:conf/nips/JeongLKK19, DBLP:conf/wacv/TangRWXX21} and pseudo-label generation~\cite {DBLP:conf/iccv/Xu00WWWB021, DBLP:conf/cvpr/WangYZ0L18}. 
The challenge increases, however, since here not only the respective concept must be correctly assigned, but also its position in the image must be detected.

% TODO: Art retrieval

\section{Semi-supervised Pose Estimation}
\label{chp:method}

In this section, we describe our method for automatic domain adaptation for human pose estimation.
First, we introduce the two-stage Transformer-based detection model in Section~\ref{sec:detection}. We then use it in the common approach of fine-tuning pre-trained models with stylized, approximately domain-specific images.
%, typically using style transfer techniques.
%As a baseline, we use the common approach of fine-tuning pre-trained models with stylized, approximately domain-specific images, typically using style transfer techniques. 
In Section~\ref{sec:semi}, we demonstrate how \enquote*{real} art-historical images can be used in the training stages with the extension of a semi-supervised process.
%We also demonstrate how \enquote*{real} art-historical images can be used in the training stages.

\begin{figure*}
\begin{center}
\includegraphics[width=.9\linewidth]{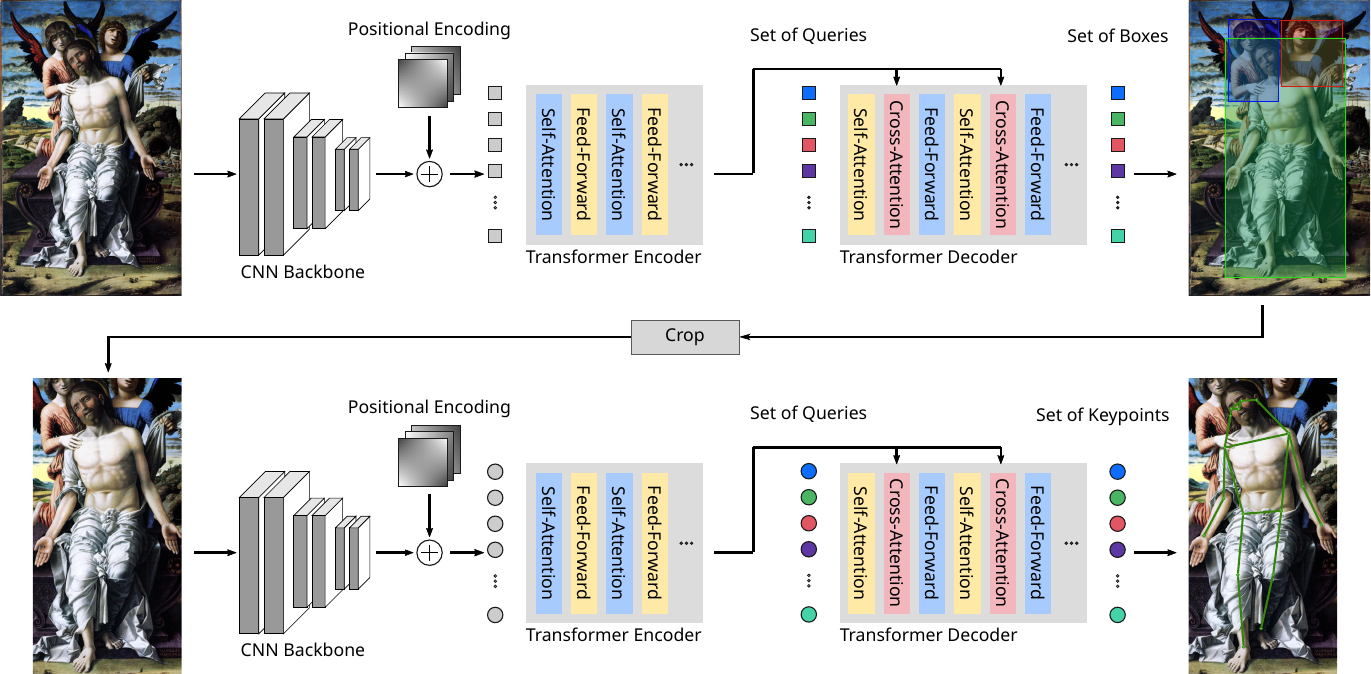}
\end{center}
\caption{Two-stage human pose estimator with Transformer~\cite{DBLP:conf/nips/VaswaniSPUJGKP17, DBLP:conf/eccv/CarionMSUKZ20}. 
The input of the first model is the entire image, which, using a \acf{CNN} backend and appropriate positional encoding, serves as input to a Transformer that predicts a fixed set of person bounding boxes. 
After filtering irrelevant detections, the individual boxes are cropped and serve as input for the second stage. 
This second Transformer model computes a set of keypoints that serve as the final prediction after filtering background classes.}
\label{fig:two_stages}
\end{figure*}

\subsection{Transformer-based Detection}\label{sec:detection}

The proposed approach is organized in two steps: 
first, persons are detected in an input image and bounding boxes are computed; in a second step, the individual boxes are scanned for keypoints. 
The initial system is based on Li et al.'s method~\cite{DBLP:conf/cvpr/0012WZXXT21}, which is built on two Transformer models for object detection~\cite{DBLP:conf/nips/VaswaniSPUJGKP17, DBLP:conf/eccv/CarionMSUKZ20}. 
The overall architecture is shown in Figure~\ref{fig:two_stages}.

In the \textbf{person detection phase}, feature descriptors are computed using a \ac{CNN} backend combined with a two-dimensional position embedding. 
After this input is flattened into a sequence of visual features, it is passed to a Transformer encoder, which is later used in the cross-attention modules of the decoder. 
The other input of the Transformer decoder is a fixed set of trainable query embeddings, where the size of the set represents the maximum number of objects to be detected. 
The output is fed into two \ac{MLP} heads. 
The first head acts as a classifier and distinguishes between person $c_{b,i}$ and background $\emptyset$, while the second one performs a regression on four %$4$ 
outputs for the position and size of the corresponding box $b_i \in [0,1]^4$. 
At the beginning of the \textbf{keypoint prediction stage}, visual features for each bounding box are determined using a \ac{CNN} backend. 
The image features, combined with position encoding and a new set of input query embeddings, are transformed to a fixed set of keypoint predictions using the Transformer. 
The main difference between the two models is that the prediction head predicts only the coordinates of keypoints $k_i \in [0,1]^2$, %, thus regressing the vectors of two %$2$ 
%elements
and instead of predicting only the person or background, classifies the type of keypoint $c_{k,i}$.

During the training phase, it is necessary to match the fixed set of predictions with the variable number of ground-truth labels per image. We thus need to find an optimal assignment $\hat{\sigma}$ between prediction $\hat{y}$ and ground-truth labels $y$ in the permutation of $N$ elements $\sigma \in \mathfrak{S}_N$ with the lowest matching cost $L_m$:
\begin{align}
    \hat{\sigma}&= \argmin_{\sigma \in \mathfrak{S}_N}\sum_{i}^{N}L_m\left(y_i,\hat{y}_{\sigma\left(i\right)}\right)
\end{align}
The optimal solution for this problem can be solved using the Hungarian algorithm~\cite{kuhn1955hungarian} and yields the assignment function $\hat{\sigma}\left(i\right)$. The assignment loss includes both the class probability and the position of the predicted object compared to the ground-truth annotation. For bounding box prediction with index $\sigma(i)$, we define the class probability $c_{b,i}$ as $\hat{p}_{\sigma\left(i\right)}\left(c_{b,i}\right)$ and the predicted box as $\hat{b}_{\sigma\left(i\right)}$. Similarly, for keypoint prediction, we define the probability of class $c_{k,i}$ as $\hat{p}_{\sigma\left(c_{k,i}\right)}\left(i\right)$ and the predicted keypoint as $\hat{k}_{\sigma\left(i\right)}$.
With these definitions, we establish the following loss functions:
% In order to compute the  We define the probability $c_{b,i}$ as $\hat{p}_{\sigma\left(i\right)}\left(c_{b,i}\right)$of the keypoint class $c_{k,i}$ as $\hat{p}_{\sigma\left(i\right)}\left(c_{k,i}\right)$.
\begin{align}
    L_{m, b}\left(y,\hat{y}\right)&= -\mathbbm{1}_{\left\{c_{b,i}\neq \emptyset\right\}}\hat{p}_{\sigma\left(i\right)}\left(c_{b,i}\right)+\mathbbm{1}_{\left\{c_{b,i}\neq\emptyset\right\}}L_b\left(b_i,\hat{b}_{\sigma\left(i\right)}\right)\\
    L_{m, k}\left(y,\hat{y}\right)&= -\mathbbm{1}_{\left\{c_{k,i}\neq \emptyset\right\}}\hat{p}_{\sigma\left(i\right)}\left(c_{k,i}\right)+\mathbbm{1}_{\left\{c_{k,i}\neq\emptyset\right\}}L_k\left(k_i,\hat{k}_{\sigma\left(i\right)}\right)
\end{align}
For bounding box prediction, the class probability defined as the $L1$-distance of the bounding box $b_i$, and the \ac{GIoU}~\cite{DBLP:conf/cvpr/RezatofighiTGS019} $L_{iou}\left(\cdot,\cdot\right)$ are chosen as the basis for cost function $L_b$. 
For keypoints $k_i$, only the class probability and the $L1$-distance of the coordinates are considered:

\begin{align}
L_b\left(b_i,\hat{b}_{\sigma\left(i\right)}\right)&=\lambda_{iou}L_{iou}\left(b_i,\hat{b}_{\sigma\left(i\right)}\right)+\lambda_{L1}\left\|b_i-\hat{b}_{\sigma\left(i\right)}\right\|\\
L_k\left(k_i,\hat{k}_{\sigma\left(i\right)}\right)&=\lambda_{L1}\left\|k_i-\hat{k}_{\sigma\left(i\right)}\right\|
\end{align}
where hyperparameters $\lambda_{iou}$ and $\lambda_{L1}$ indicate the weight of each loss component. 
Predictions that could not be assigned to a ground-truth label are instead assigned to the background class $\emptyset$ during optimization; 
their bounding boxes and keypoint coordinates are not considered in the loss. 
After the best assignment is found, the loss can be calculated as follows:

\begin{align}
L_{H,b}\left(y,\hat{y}\right) &= \sum_{i=1}^{N}\left[-\log\hat{p}_{\hat{\sigma}}\left(c_{b,i}\right)+\mathbbm{1}_{\left\{c_{b,i}\neq\emptyset\right\}}L_{b}\left(b_i,\hat{b}_{\hat{\sigma}}\left(i\right)\right)\right]\\
L_{H,k}\left(y,\hat{y}\right) &= \sum_{i=1}^{N}\left[-\log\hat{p}_{\hat{\sigma}}\left(c_{k,i}\right)+\mathbbm{1}_{\left\{c_{k,i}\neq\emptyset\right\}}L_{k}\left(b_i,\hat{b}_{\hat{\sigma}}\left(i\right)\right)\right]
\end{align}
During the inference of bounding box prediction, it is sufficient to filter the predicted boxes using a threshold function. However, during the inference step of keypoint prediction, it is necessary to find an optimal assignment again because the Transformer model predicts up to $N$ points, but the number is usually larger than the maximum number of possible keypoints per person. Since no ground-truth information is known during inference, the following cost function is used:
\begin{align}
    L_{m, k}\left(y,\hat{y}\right)&= -\hat{p}_{\sigma\left(i\right)}\left(c_{k,i}\right)
\end{align}
% TODO maybe something about how we test this
% TODO maybe move this part or delete it if we miss space at the end
Compared to object detection methods such as Faster Region-based CNN (R-CNN)~\cite{DBLP:conf/nips/RenHGS15} and YOLO (You Only Look Once)~\cite{DBLP:conf/cvpr/RedmonF17}, the approach does not predict multiple bounding box candidates for each image region, but only a fixed set of boxes for each image. 
This greatly simplifies post-processing, as no overlapping bounding boxes are predicted for same-person instances, and the imbalance between background and foreground classes is much smaller.

\begin{figure}
\begin{center}
\includegraphics[width=\linewidth]{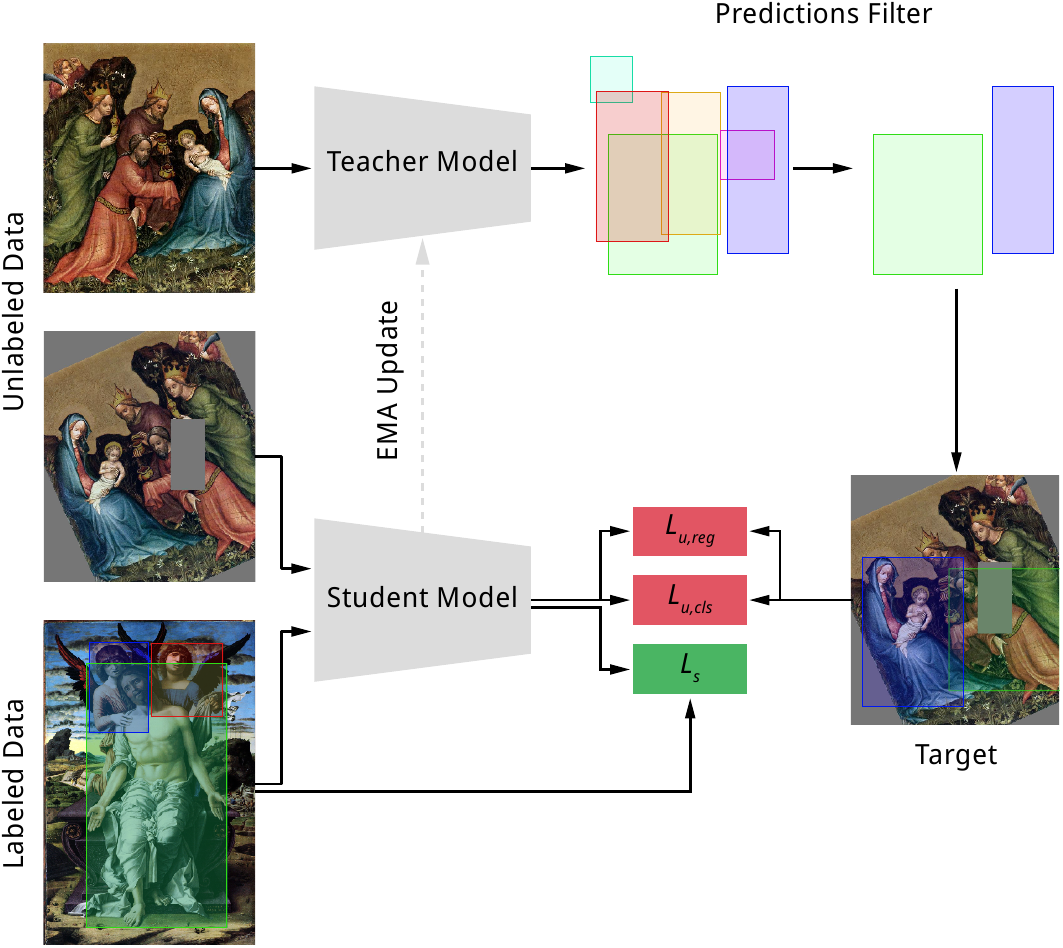}
\end{center}
\caption{The semi-supervised training pipeline adapted from \citet{DBLP:conf/iccv/Xu00WWWB021} is shown. During training, each batch consists of labeled and unlabeled images with strong and weak augmentations generated for unlabeled ones. The teacher uses the weakly labeled data to generate pseudo bounding boxes or pseudo keypoints that are used to train the strongly augmented images. This involves thresholding the predictions and then transferring the corresponding boxes or keypoints to the coordinate system of the strongly augmented image.}
\label{fig:semi}
\end{figure}

\subsection{Semi-supervised Domain Adaptation}\label{sec:semi}

To extend the available data sets for bounding box and keypoint detection in art-historical images, we augment the training pipeline by adapting the semi-supervised approach from \citet{DBLP:conf/iccv/Xu00WWWB021}. 
Since we use a Transformer model instead of a Faster R-CNN, the number of predicted bounding boxes and keypoints is considerably smaller and simplifies certain steps. 
An overview of the semi-supervised pipeline is shown in Figure~\ref{fig:semi}. 
The basic principle is to use both labeled and unlabeled examples to train a student model. 
Here, the teacher, whose weights are based on the \ac{EMA} of the student weights, serves as a generator of pseudo-labels for bounding boxes and keypoints. 
For this purpose, weakly augmented unlabeled images are used for person detection and weakly augmented cropped bounding boxes for keypoint prediction. 
Subsequently, the predicted objects are filtered with the threshold $\tau=0.9$ and projected onto the strongly augmented unlabeled images. 
Contrary to \citet{DBLP:conf/iccv/Xu00WWWB021}, we use the teacher prediction for bounding boxes and keypoints only if it is not a background class. 
In order not to distort the ratio between negative and positive boxes or keypoints, we use the same threshold to filter negative examples; but this time from the forward step of the student. 
This is necessary because there is no relationship between the predicted coordinates of the teacher's negative classes and the student's negative predictions. 
The total loss now includes a supervised component $L_s$ and an unsupervised component $L_u$. 
It is calculated as follows:
\begin{align}
    L &= L_s + \lambda_{u} L_u
\end{align}
Depending on the current target, the supervised loss is the same as for supervised learning, $L_s \in \left\{L_{H,b},L_{H,k}\right\}$. For the unsupervised loss part, we use the prediction of the teacher model to detect bounding boxes or keypoints. Therefore, for the prediction of the bounding box with index $i$, we define the probability of class $c_{b,i}$ as $\hat{p}^t\left(c_{b,i}\right)$ and the predicted box as $\hat{b}^t$. Similarly, for the teacher keypoint prediction, we define the probability of class $c_{k,i}$ as $\hat{p}^t\left(c_{k,i}\right)$ and the predicted keypoint as $\hat{k}^t$. With these definitions, we can establish the loss functions:
\begin{align}
    L_{u,reg,b}&=\sum_{i}^N\mathbbm{1}_{\left\{c_{b,i}\neq \emptyset;\, \hat{p}^t\left(c_{b,i}\right)\geq\tau\right\}}L_{b}\left(\hat{b}_i^t,\hat{b}_{\hat{\sigma}}\left(i\right)\right)\\
    L_{u,reg,k}&=\sum_{i}^N\mathbbm{1}_{\left\{c_{k,i}\neq \emptyset;\, \hat{p}^t\left(c_{k,i}\right)\geq\tau\right\}}L_{k}\left(\hat{k}_i^t,\hat{k}_{\hat{\sigma}}\left(i\right)\right)
\end{align}
The classification loss of the unlabeled examples is given by the positive classes resulting from the teacher's probability of exceeding threshold $\tau$ and the negative examples from the student's prediction. It is defined as follows:
\begin{align}
    L_{u,cls,b}=&-\sum_{i}^N
    \mathbbm{1}_{\left\{c_{b,i}\neq \emptyset;\, \hat{p}^t\left(c_{b,i}\right)\geq\tau\right\}}\log\hat{p}_{\hat{\sigma}}\left(c_{b,i}\right)\\
    &-\sum_{i}^N\mathbbm{1}_{\left\{c_{b,i}=\emptyset;\, \hat{p}_{\hat{\sigma}}\left(c_{b,i}\right)\geq\tau\right\}}\log\hat{p}_{\hat{\sigma}}\left(c_{b,i}\right)\nonumber\\
    L_{u,cls,k}=&-\sum_{i}^N
    \mathbbm{1}_{\left\{c_{k,i}\neq \emptyset;\, \hat{p}^t\left(c_{k,i}\right)\geq\tau\right\}}\log\hat{p}_{\hat{\sigma}}\left(c_{k,i}\right)\\
    &-\sum_{i}^N\mathbbm{1}_{\left\{c_{k,i}=\emptyset;\, \hat{p}_{\hat{\sigma}}\left(c_{k,i}\right)\geq\tau\right\}}\log\hat{p}_{\hat{\sigma}}\left(c_{k,i}\right)\nonumber
\end{align}

\section{Experimental Setup and Results}
\label{chp:exp}

In this section, we introduce our data sets as well as discuss the conducted quantitative and qualitative studies. For the training and test phases of our pipelines, we use various real-world, synthetically generated, and art-historical data sets (Section~\ref{chp:datasets}). To evaluate the performance of each model and approach, we first conduct a series of ablation studies (Section~\ref{chp:ablation}) and then qualitatively assess our method's ability to provide reasonable predictions (Section~\ref{chp:qualitative}). To evaluate the experiments, we use the metrics\footnote{\url{https://cocodataset.org/\#keypoints-eval}, last accessed on \today.} and tools provided by the COCO API.\footnote{\url{https://github.com/cocodataset/cocoapi}, last accessed on \today.}

\begin{table}[]
\caption{Overview of the data sets used in our experiments. Persons are indicated by bounding boxes associated with them. Up to 17 keypoints are stored per person.}
\label{tab:data}

\begin{tabularx}{\columnwidth}{@{}Xlrrr@{}}
\toprule
Data set      & Split        & Images  & Persons & Keypoints \\
\midrule
COCO 2017     & Training     & 118,287 & 262,465 & 1,642,283 \\
              & Validation   & 5,000   & 11,004  & 68,215    \\
              & Test         & 0       & 0       & 0         \\
              \cmidrule{2-5}
              & Total        & 123,287 & 273,469 & 1,710,498 \\
\midrule
COCO 2017     & Training     & 236,574 & 524,930 & 3,284,566 \\
(stylized)    & Validation   & 10,000  & 22,008  & 136,430   \\
              & Test         & 0       & 0       & 0         \\
              \cmidrule{2-5}
              & Total        & 246,574 & 546,938 & 3,420,996 \\
\midrule
People-Art    & Training     & 1,746   & 1,330   & 0         \\
              & Validation   & 1,489   & 1,080   & 0         \\
              & Test         & 1,616   & 1,088   & 0         \\
              \cmidrule{2-5}
              & Total        & 4,851   & 3,498   & 0         \\
\midrule
PoPArt        & Training     & 1,553   & 2,069   & 30,415    \\
              & Validation   & 643     & 704     & 10,367    \\
              & Test         & 663     & 741     & 10,863    \\
              \cmidrule{2-5}
              & Total        & 2,859   & 3,514   & 51,645    \\
\midrule
ART500k       & Training     & 318,869 & 0       & 0         \\
              & Validation   & 0       & 0       & 0         \\
              & Test         & 0       & 0       & 0         \\
              \cmidrule{2-5}
              & Total        & 318,869 & 0       & 0         \\
\bottomrule
\end{tabularx}
\end{table}

\subsection{Data Sets}
\label{chp:datasets}

An overview of the data sets used in our experiments with their respective splits is shown in Table~\ref{tab:data}. 
All data sets are based on the \ac{COCO} format, where each person instance is labeled with up to $17$ keypoints.

The largest annotated data set results from the \textbf{COCO 2017} detection and keypoint challenge, which includes $118,287$ training and $5,000$ validation images with person instances.\footnote{\url{https://www.kaggle.com/datasets/awsaf49/coco-2017-dataset}, last accessed on \today.} 
To evaluate the performance of the common scenario that uses style transfer to close the domain gap between annotated real-world training and art-historical inference data, we generate a stylized version of the data set. 
For this purpose, we leverage the style transfer approach from~\citet{chen2021artistic} to create two style variants for each \ac{COCO} image, where the style images are randomly selected from the Painter by Numbers data set~\cite{painter}.

The models are grounded in two domain-specific, sufficiently large data sets that recycle openly licensed subsets of the art-historical online encyclopedia WikiArt\footnote{\url{https://www.wikiart.org/}, last accessed on \today}: the 2016 compiled People-Art data set~\cite{DBLP:journals/corr/CaiWCH15, DBLP:conf/eccv/WestlakeCH16}, in which human figures are marked with bounding boxes enclosing them. 
The second data set, called \ac{PoPArt}, is introduced here and identifies $17$ limb points in addition to bounding boxes. 
Both data sets approximately reflect the diversity of art-historical depictions of human figures through time by featuring $43$ and $22$ different styles, respectively; ranging from impressionistic to neo-figurative and realistic variants. 
The pre-existing \textbf{People-Art} data set is enhanced on two levels. 
First, we integrate additional negative examples of mammals that were frequently false positively classified as humans~\cite{DBLP:conf/eccv/WestlakeCH16}. 
Second, we use the largest resolution of images provided by WikiArt to avoid further complicating the detection of relatively small figures due to possible image artifacts in low-resolution reproductions. After these preparatory measures, People-Art features $1,746$ training, $1,489$ validation, and $1,616$ test images. 
The annotation of the novel \textbf{\ac{PoPArt}} data set was performed according to the following principles (see Figure~\ref{fig:examples-gt} for some examples with ground-truth annotations): 
(1)~the body of a human figure must be recognizable, which implies that more than six keypoints are annotatable, covering at least head and shoulder area;
(2)~a maximum of four figures are annotated per image; if more than four instances are shown, those whose body permits to annotate as many limbs as possible are selected;
(3)~if an occluded body part can be sufficiently approximated by another visible one, the respective associated keypoint is annotated;
(4)~in profile views, eyes and ears are usually annotated on the non-visible side of the face as well. 
The data set includes $1,553$ training, $643$ validation, and $663$ test images, where each split contains proportionally the same number of images per style.

With the \textbf{ART500k} data set~\cite{DBLP:conf/mm/MaoCS17}, we moreover integrate an art-historical data set not annotated with person instances into the training procedure. A {\sisetup{round-precision=0}\SI{50}{\percent}} split of all ART500k images with a total of $318,869$ examples is generated, which we use in our semi-supervised learning approach as unlabeled data.

%\begin{figure}
%\centering
%\begin{subfigure}{.495\columnwidth}
%  \centering
%  \includegraphics[width=\linewidth]{images/bronzino.png}
%  \caption{}
%\end{subfigure}%
%\hfill
%\begin{subfigure}{.495\columnwidth}
%  \centering
%  \includegraphics[width=\linewidth]{images/bronzino_pose-estimation.png}
%  \caption{}
%\end{subfigure}
%\caption{Agnolo Bronzino's \textit{Venus, Cupid, and Jealousy} (c. 1549) without (a) and with (b) ground-truth bounding box and keypoint annotations for Cupid (red) and Venus (green), respectively.}
%\label{fig:bronzino}
%\end{figure}

\subsection{Ablation Study}
\label{chp:ablation}

\label{chp:exp_boxes}
For \textbf{person detection}, we leverage the weights of a \ac{DETR} model~\cite{DBLP:conf/eccv/CarionMSUKZ20} pre-trained on COCO 2017 and reinitialize the classification head. 
An Adam optimizer~\cite{DBLP:journals/corr/KingmaB14} with a learning rate of $lr = 5e-6$ is used for the Transformer and with $lr = 1e-7$ for the ResNet-50 backbone~\cite{DBLP:conf/cvpr/HeZRS16}. Similar to \citet{DBLP:conf/cvpr/0012WZXXT21}, all classes except persons are ignored; small bounding boxes are not considered. 
Models are trained for $200,000$ iterations with a batch size of four, with all images randomly scaled to a maximum size of $1,333$ pixels per side. When training the semi-supervised models, the batch size is increased by four additional unlabeled images. 
The weights of the different loss hyperparameters are set to $\lambda_{L1} = 5$, $\lambda_{iou} = 2$, and $\lambda_u = 0.5$.

\begin{table*}
\caption{Person detection results on the People-Art and PoPArt test sets, respectively. For PoPArt, $AP_{S}$ is neglected as no test data is available for small human figures, most of which have no annotatable pose due to their size. The best performing approach per test set is bold.}
\label{tab:exp_boxes}

\begin{tabularx}{\textwidth}{@{}XXRRRRRRRRR@{}}
\toprule
Test set & Train set & Stylized & Semi & $AP$ & $AP_{50}$ & $AP_{75}$ & $AP_{S}$ & $AP_{M}$ & $AP_{L}$ & $AR$  \\
\midrule
People-Art & COCO 2017 & {\sisetup{round-precision=0}\SI{0}{\percent}}&&\num{0.3118} & \num{0.5106} & \num{0.3175} & \num{0.0075} & \num{0.2118} & \num{0.3294} & \num{0.6728} \\
& COCO 2017 & {\sisetup{round-precision=0}\SI{0}{\percent}}&\checkmark&\num{0.3696} & \num{0.597} & \num{0.3885} & \num{0.0007} & \num{0.2115} & \num{0.395} & \textbf{\num{0.7351}} \\
& COCO 2017 & {\sisetup{round-precision=0}\SI{50}{\percent}}&&\num{0.3686} & \num{0.6113} & \num{0.3871} & \num{0.0045} & \num{0.2386} & \num{0.3941} & \num{0.7257} \\
& COCO 2017 & {\sisetup{round-precision=0}\SI{50}{\percent}}&\checkmark&\num{0.3744} & \num{0.6277} & \num{0.3792} & \num{0.0024} & \num{0.2193} & \num{0.4011} & \num{0.7296} \\
& COCO 2017 & {\sisetup{round-precision=0}\SI{100}{\percent}}&&\num{0.3727} & \num{0.6256} & \num{0.3922} & \num{0.024} & \num{0.2406} & \num{0.3981} & \num{0.7165} \\
& COCO 2017 & {\sisetup{round-precision=0}\SI{100}{\percent}}&\checkmark&\num{0.3846} & \num{0.6333} & \num{0.4047} & \num{0.0115} & \num{0.2313} & \num{0.4108} & \num{0.7221} \\
% & PoPArt & {\sisetup{round-precision=0}\SI{0}{\percent}}&&\num{0.3441} & \num{0.5463} & \num{0.3615} & \num{0.0609} & \num{0.1411} & \num{0.3744} & \num{0.6999} \\
% & PoPArt & {\sisetup{round-precision=0}\SI{0}{\percent}}&\checkmark&\num{0.3662} & \num{0.5755} & \num{0.3886} & \num{0.0676} & \num{0.1658} & \num{0.3975} & \num{0.7248} \\
\cmidrule{2-11}
& People-Art & {\sisetup{round-precision=0}\SI{0}{\percent}} & & \num{0.428} & \num{0.7279} & \num{0.435} & \textbf{\num{0.0676}} & \num{0.2123} & \num{0.4636} & \num{0.7041} \\
& People-Art & {\sisetup{round-precision=0}\SI{0}{\percent}}&\checkmark&\textbf{\num{0.4428}} & \textbf{\num{0.7381}} & \textbf{\num{0.459}} & \num{0.0509} & \textbf{\num{0.2412}} & \textbf{\num{0.4769}} & \num{0.7291} \\
\midrule
PoPArt & COCO 2017 & {\sisetup{round-precision=0}\SI{0}{\percent}}&&\num{0.2287} & \num{0.3041} & \num{0.2433} & & \num{0.1096} & \num{0.2336} & \num{0.7997} \\
& COCO 2017 & {\sisetup{round-precision=0}\SI{0}{\percent}}&\checkmark&\num{0.2422} & \num{0.3353} & \num{0.2612} & & \num{0.0324} & \num{0.2469} & \num{0.8377} \\
& COCO 2017 & {\sisetup{round-precision=0}\SI{50}{\percent}}&&\num{0.2322} & \num{0.3168} & \num{0.248} & & \num{0.04} & \num{0.2397} & \num{0.8365} \\
& COCO 2017 & {\sisetup{round-precision=0}\SI{50}{\percent}}&\checkmark&\num{0.2261} & \num{0.3125} & \num{0.2452} & & \num{0.0347} & \num{0.2324} & \num{0.8277} \\
& COCO 2017 & {\sisetup{round-precision=0}\SI{100}{\percent}} & & \num{0.2542} & \num{0.354} & \num{0.273} & & \num{0.036} & \num{0.2624} & \num{0.8128} \\
& COCO 2017 & {\sisetup{round-precision=0}\SI{100}{\percent}} & \checkmark & \num{0.2359} & \num{0.331} & \num{0.2516} & & \num{0.048} & \num{0.2423} & \num{0.8284} \\
% & People-Art &{\sisetup{round-precision=0}\SI{0}{\percent}}&&\num{0.2585} & \num{0.3728} & \num{0.2874} & \num{0.0621} & \num{0.2663} & \num{0.7976} \\
% & People-Art & {\sisetup{round-precision=0}\SI{0}{\percent}}&\checkmark&\num{0.2612} & \num{0.3695} & \num{0.289} & \num{0.0431} & \num{0.2682} & \num{0.82} \\
\cmidrule{2-11}
& PoPArt & {\sisetup{round-precision=0}\SI{0}{\percent}} & & \num{0.4898} & \num{0.6566} & \num{0.5279} & & \textbf{\num{0.2639}} & \num{0.4945} & \num{0.8468} \\
& PoPArt & {\sisetup{round-precision=0}\SI{0}{\percent}} & \checkmark & \textbf{\num{0.5073}} & \textbf{\num{0.6728}} & \textbf{\num{0.5302}} & & \num{0.2132} & \textbf{\num{0.5119}} & \textbf{\num{0.8561}} \\
\bottomrule
\end{tabularx}
\end{table*}

The results for the respective test sets are shown in Table~\ref{tab:exp_boxes}. 
We notice that our semi-supervised learning technique on People-Art always results in an improvement of \ac{AP} and \ac{AR}. 
Moreover, \ac{AP} maintains this advantage as the proportion of style-transferred material increases, but becomes successively smaller. 
The domain-specific data further increases the performance significantly, such that \ac{AP} rises from \num{0.428} to \num{0.4428} and \ac{AR} from \num{0.7041} to \num{0.7291}. 
With $AP_{50} = 0.7381$, the performance of our approach is considerably above the best results of $AP_{50} = 0.68$ and $AP_{50} = 0.583$ reported so far by \citet{DBLP:conf/ijcnn/KadishRL21} and \citet{DBLP:journals/cviu/GonthierLG22} for the data set, respectively. 
For \ac{PoPArt}, we find that semi-supervised learning with art-historical images enhances \ac{AP} less; thus, our proposed method with COCO 2017 annotations has similar performance to using style transfer. 
The comparison between training with COCO 2017 data and training on \ac{PoPArt} indicates a larger improvement especially in \ac{AP}. 
This deviation can be explained by the different types of annotations, as \ac{PoPArt} was annotated exclusively for pose estimation and contains fundamentally fewer ground-truth bounding boxes of human figures. 
Nevertheless, our proposed semi-supervised learning approach is beneficial: the performance increases from \num{0.4898} to \num{0.5073} for \ac{AP} and from \num{0.8468} to \num{0.8561} for \ac{AR}.

\label{chp:exp_keypoint}
In the \textbf{keypoint prediction} stage we use the \acl{HRNet} with 32 feature channels (HRNet-W32) as backbone with an input resolution of $384 \times 288$ pixels~\cite{DBLP:conf/cvpr/0009XLW19}. 
Again, we leverage the pre-trained weights on COCO 2017 provided by \citet{DBLP:conf/cvpr/0012WZXXT21} and reinitialize the classification layer. 
The model is trained for $150,000$ iterations with a batch size of $16$; the learning rates are set to $lr = 1e-5$ for the Transformer and $lr = 1e-6$ for the HRNet. 
We then divide both rates by $10$ and train for another $50,000$ iterations with Adam. 
For the semi-supervised methods, we add to the batch $16$ unlabeled images generated from the models' predictions from Table~\ref{tab:exp_boxes} on ART500k. 
Predicted bounding boxes whose confidence level is above $0.5$ are used for this purpose. 
The effects of keypoint prediction are similar to those of person detection: 
we observe that \ac{AR} can be significantly improved by our semi-supervised learning technique. 
Models not only trained with %more %than 
style-transferred images also show an increase in \ac{AP}. 
In particular, for those using \ac{PoPArt}, \ac{AP} rises from \num{0.4844} to \num{0.5258} and \ac{AR} from \num{0.7078} to \num{0.7464}. 
Results for the \ac{PoPArt} test set are summarized in Table~\ref{tab:exp_keypoints} with OpenPose~\cite{DBLP:journals/pami/CaoHSWS21} included for additional reference.

\begin{table*}
\caption{Keypoint detection results on the PoPArt test set with predicted bounding boxes of the model with the same strategy. 
$AP_{S}$ is neglected as no test data is available for small human figures, most of which have no annotatable pose due to their size. 
For PoPArt train sets, the first entry refers to the training data set used for bounding box detection and the second to the training data set used for keypoint prediction. 
The best performing approach is bold.}
\label{tab:exp_keypoints}

\begin{tabularx}{\textwidth}{@{}XXRRRRRRRR@{}}
\toprule
Test set & \multicolumn{3}{l}{Method} & $AP$ & $AP_{50}$ & $AP_{75}$ & $AP_{M}$ & $AP_{L}$ & $AR$  \\
\midrule
PoPArt & \multicolumn{3}{l}{OpenPose} & \num{0.1388} & \num{0.2534} & \num{0.1283} & \num{0.028} & \num{0.1417} & \num{0.4382} \\
\cmidrule{2-10}
& Train set & Stylized & Semi & $AP$ & $AP_{50}$ & $AP_{75}$ & $AP_{M}$ &$AP_{L}$ & $AR$  \\
\cmidrule{2-10}
& COCO 2017 & {\sisetup{round-precision=0}\SI{0}{\percent}} & & \num{0.2285} & \num{0.2811} & \num{0.2545} & \num{0.0236} & \num{0.2367} & \num{0.554} \\
\phantom{People-Art} & COCO 2017 & {\sisetup{round-precision=0}\SI{0}{\percent}} & \checkmark & \num{0.2525} & \num{0.3173} & \num{0.281} & \num{0.0122} & \num{0.2639} & \num{0.7009} \\
& COCO 2017 & {\sisetup{round-precision=0}\SI{50}{\percent}} & & \num{0.2401} & \num{0.3072} & \num{0.2672} & \num{0.0215} & \num{0.2531} & \num{0.6672} \\
& COCO 2017 & {\sisetup{round-precision=0}\SI{50}{\percent}} & \checkmark & \num{0.2413} & \num{0.3052} & \num{0.2665} & \num{0.018} & \num{0.2554} & \num{0.688} \\
& COCO 2017 & {\sisetup{round-precision=0}\SI{100}{\percent}} & & \num{0.2657} & \num{0.3426} & \num{0.2932} & \num{0.0153} & \num{0.2845} & \num{0.6765} \\
& COCO 2017 & {\sisetup{round-precision=0}\SI{100}{\percent}} & \checkmark & \num{0.2518} & \num{0.3167} & \num{0.2813} & \num{0.0169} & \num{0.2653} & \num{0.6896} \\
\cmidrule{2-10}
& People-Art/PoPArt &{\sisetup{round-precision=0}\SI{0}{\percent}} & & \num{0.2841} & \num{0.3622} & \num{0.3073} & \num{0.0378} & \num{0.2916} & \num{0.7185} \\
& People-Art/PoPArt &{\sisetup{round-precision=0}\SI{0}{\percent}} & \checkmark &\num{0.2971} & \num{0.3637} & \num{0.3272} & \num{0.0204} & \num{0.3118} & \textbf{\num{0.7583}} \\
\cmidrule{2-10}
& PoPArt/PoPArt & {\sisetup{round-precision=0}\SI{0}{\percent}} & & \num{0.4844} & \num{0.606} & \num{0.5319} & \textbf{\num{0.0771}} & \num{0.492} & \num{0.7078} \\
& PoPArt/PoPArt & {\sisetup{round-precision=0}\SI{0}{\percent}} & \checkmark & \textbf{\num{0.5258}} & \textbf{\num{0.6392}} & \textbf{\num{0.5735}} & \num{0.0308} & \textbf{\num{0.535} }& \num{0.7464} \\
\bottomrule
\end{tabularx}
\end{table*}

%\begin{figure}
%\centering
%  \includegraphics[width=\linewidth]{graphics/examples.pdf}
%\caption{Comparison of the prediction of different models. The illustration includes sections from the popart test data set with the ground-truth annotation \textbf{(a)} and the predictions of Openpose \textbf{(b)}, COCO without style transfer and without semi-supervised training \textbf{(c)}, and the prediction of the model trained on \ac{PoPArt} with semi-supervised learning \textbf{(d)}.}
%\label{fig:examples}
%\end{figure}

\begin{figure}
\centering
\begin{subfigure}{.49\linewidth}
  \centering
  \includegraphics[width=\linewidth]{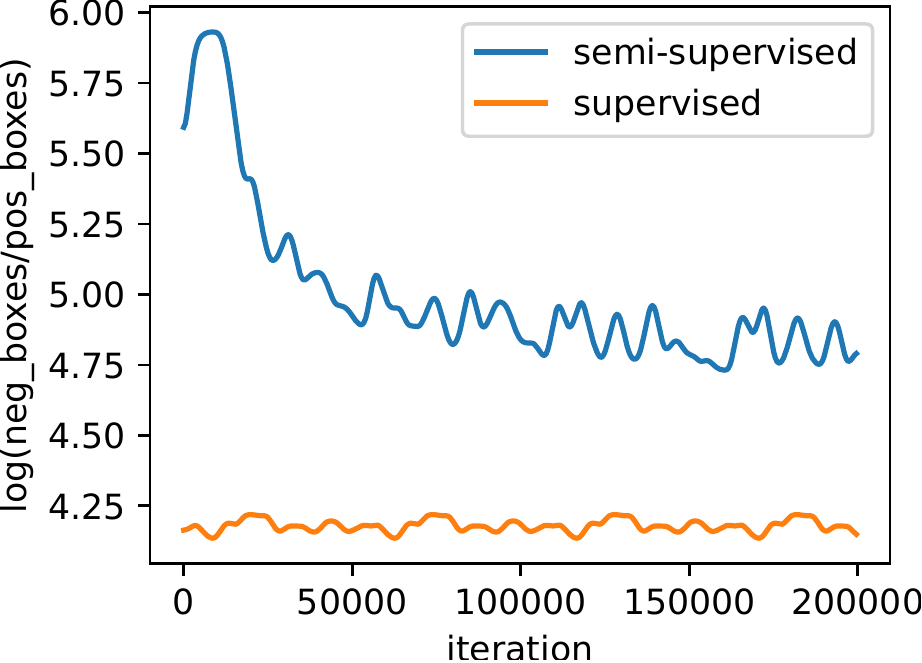}
  \caption{}
  \label{fig:distribution-boxes}
\end{subfigure}%
\hfill
\begin{subfigure}{.49\linewidth}
  \centering
  \includegraphics[width=\linewidth]{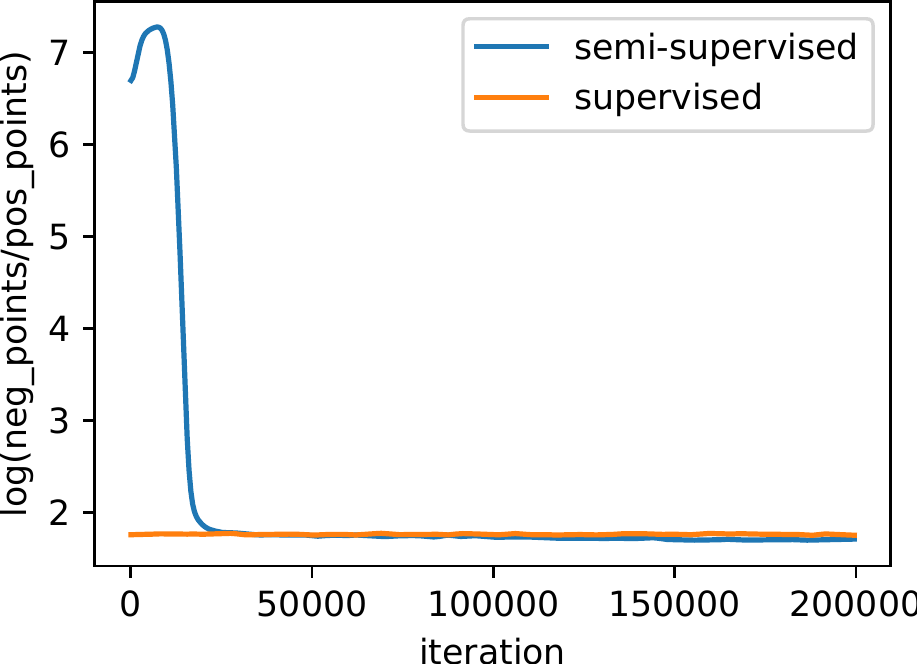}
  \caption{}
  \label{fig:distribution-keypoints}
\end{subfigure}%
\caption{Distribution of positive and negative classes on \ac{PoPArt} (orange) and the teacher's predicted distribution for unlabeled data on ART500k (blue). It is evident that the teacher recognizes fewer bounding boxes in the person detection phase (a) and estimates more points in the keypoint prediction phase (b) in comparison.}
% Distribution of positive and negative ($\emptyset$) classes given by dataset (blue) and Teacher's predicted distribution on unlabeled data (orange). The \ac{PoPArt} dataset (orange) shows that in comparison to the ART500k dataset (blue) fewer boxes are recognized by the Teacher in the bounding boxes stage and the teacher even predicts more points during the keypoints stage.
\label{fig:distribution}
\end{figure}

To evaluate the behavior of our \textbf{semi-supervised approach}, we examine the number of positive and negative teacher predictions during training. 
To this end, we illustrate the ratios between negative and positive bounding boxes (Figure~\ref{fig:distribution-boxes}) and keypoints (Figure~\ref{fig:distribution-keypoints}) of the labeled and unlabeled parts of a batch. 
As we compute the target labels for the background class directly from the student's predictions, we can see in both cases that the ratio of the background increases sharply until it reaches the maximum at about $10,000$ iterations. 
After that, it starts to decrease in favor of positive classes as the confidence score of the teacher's predictions starts to exceed threshold~$\tau$. 
In case of keypoints, it becomes apparent that the ratio between supervised (\ac{PoPArt}) and unsupervised components (ART500k) per batch is equalized, and later on average more keypoints are detected in ART500k than in \ac{PoPArt}.

\begin{figure}
\centering
\begin{subfigure}{.247\linewidth}
  \centering
  \includegraphics[width=\linewidth]{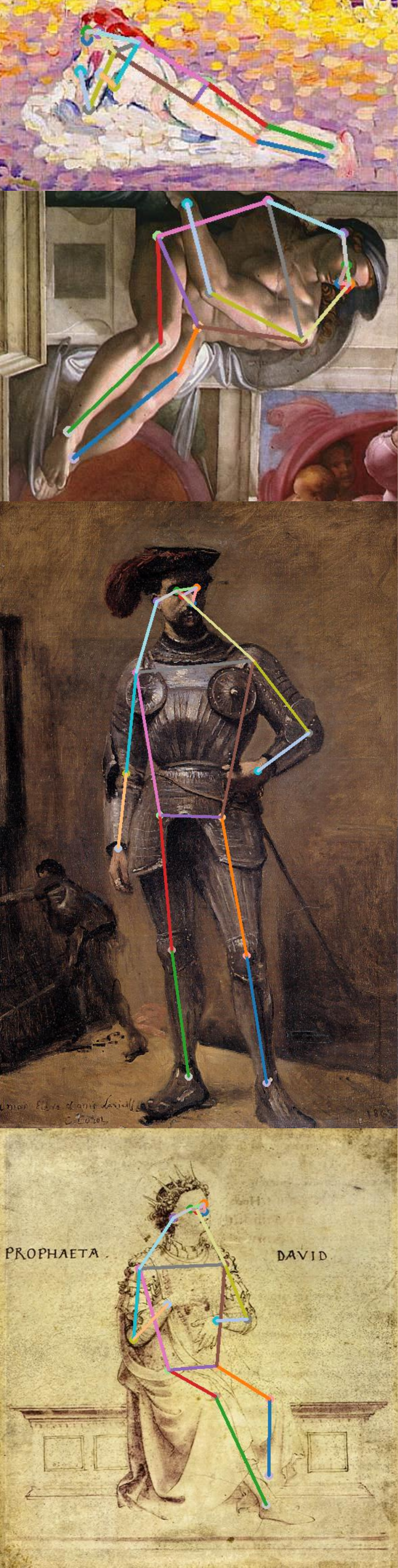}
  \caption{}
  \label{fig:examples-gt}
\end{subfigure}%
\hfill
\begin{subfigure}{.247\linewidth}
  \centering
  \includegraphics[width=\linewidth]{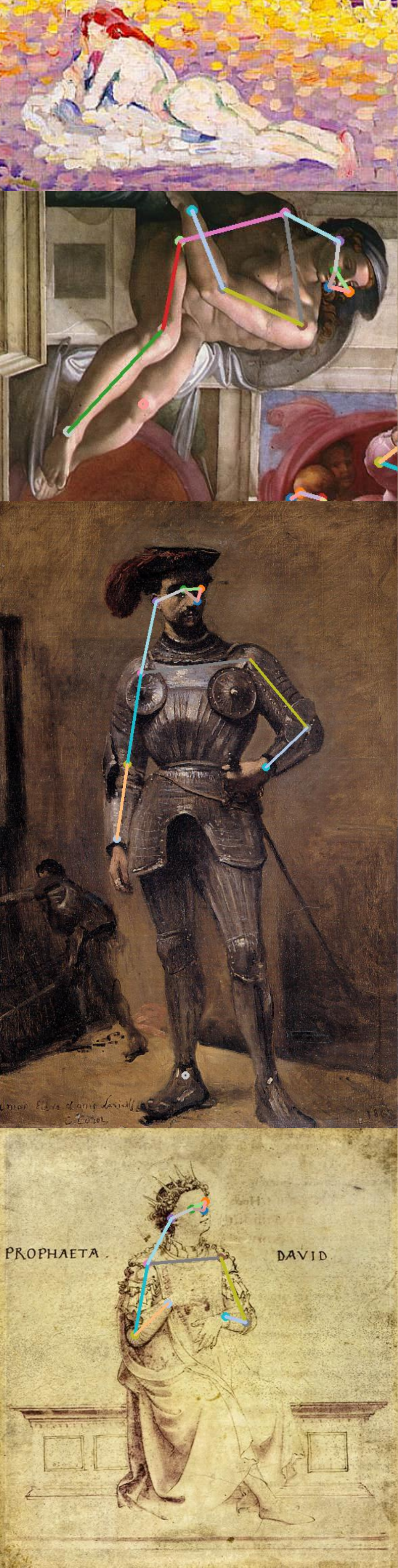}
  \caption{}
  \label{fig:examples-openpose}
\end{subfigure}%
\hfill
\begin{subfigure}{.247\linewidth}
  \centering
  \includegraphics[width=\linewidth]{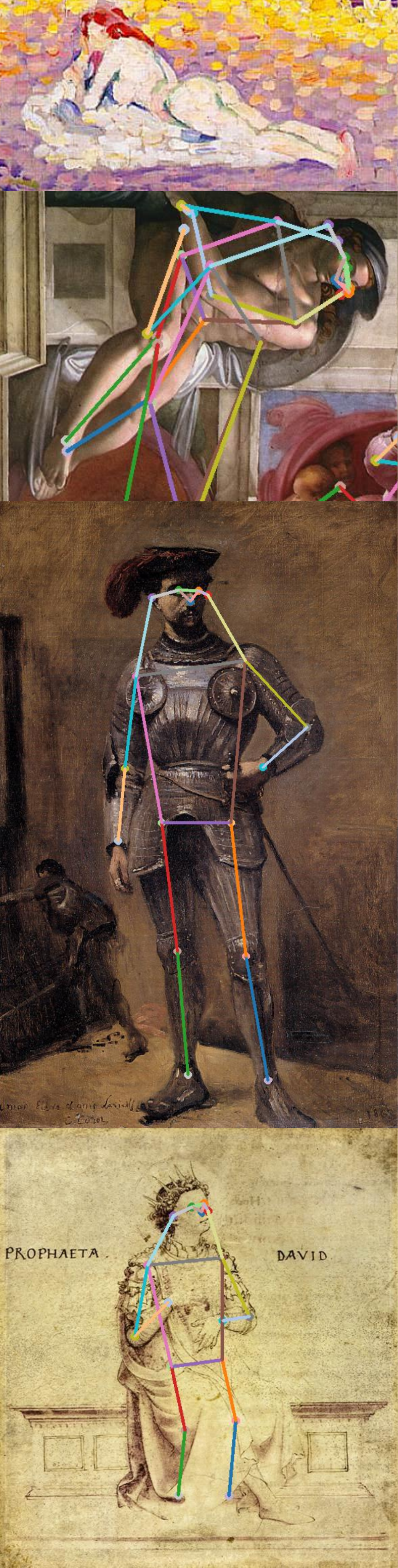}
  \caption{}
  \label{fig:examples-coco}
\end{subfigure}%
\hfill
\begin{subfigure}{.247\linewidth}
  \centering
  \includegraphics[width=\linewidth]{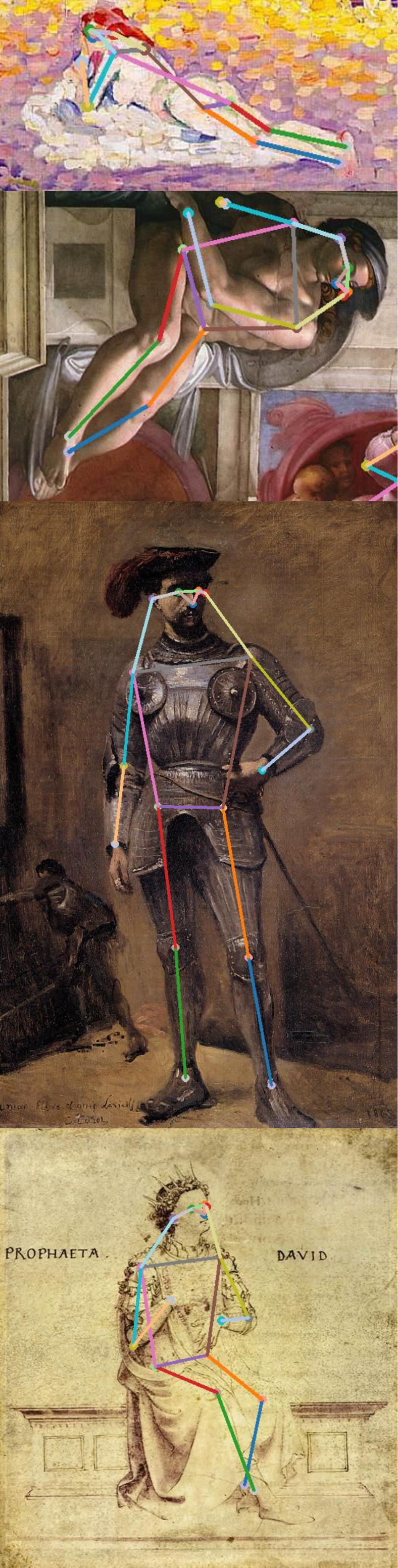}
  \caption{}
  \label{fig:examples-our}
\end{subfigure}%
\caption{Predictions of the different models with examples from the \ac{PoPArt} test data. (a) ground-truth annotations; (b) estimations of OpenPose; (c) COCO 2017 without style transfer and semi-supervised training; (d) \ac{PoPArt} trained with semi-supervised learning.}
\label{fig:examples}
\end{figure}

\subsection{Qualitative Analysis}
\label{chp:qualitative}

To qualitatively assess our method's ability to provide reasonable predictions, we visually compare it to ground-truth annotations and two of the other models. 
Figure~\ref{fig:examples-openpose} illustrates that OpenPose almost consistently tends to estimate only parts of the face and some points of the torso; holistically correct predictions are rare. 
Bodies in non-realistic settings are often not captured, exemplified by Henri Edmond Cross's Neo-Impressionist example from the early 20th century (Figure~\ref{fig:examples-openpose}, \textit{first row}). 
This is also noticeable in the model trained on COCO 2017 without style transfer and unsupervised learning (Figure~\ref{fig:examples-coco}). 
However, more suitable approximations of the lower body are identified, at least for Jean-Baptiste Camille Corot's \textit{Knight} (1868; Figure~\ref{fig:examples-coco}, \textit{third row}) and Fra Angelico's religious drawing of King David (ca. 1430; \textit{fourth row}). 
Highly problematic, though, are hidden limbs or bodies not depicted from usual perspectives, illustrated by the detail of Michelangelo's Sistine Chapel ceiling painting in Figure~\ref{fig:examples-coco} (\textit{second row}). 
Our proposed model, trained on \ac{PoPArt} and with semi-supervised learning, even manages predominantly complex scenarios (Figure~\ref{fig:examples-our}). 
Minor errors result from limbs assigned to the wrong side of the body (\textit{first row}), poorly contrasting or rather abstractly drawn body parts---or overlaps with limbs of other persons, which in \ac{PoPArt} were primarily due to Aubrey Beardsley's works. 
This is especially true for styles that introduce complications even when manually labeled, e.g., in case of the Japanese genre Ukiyo-e, since expressive poses with strongly flowing robes often lack clear assignment of joint points. 
%In addition, the correct assignment of points can be disturbed in content showing a person with his or her mirror image.
In addition, the correct assignment of points can be disturbed if the image shows a person and his or her mirror image.

\begin{figure}
\centering 
\includegraphics[width=0.96\linewidth]{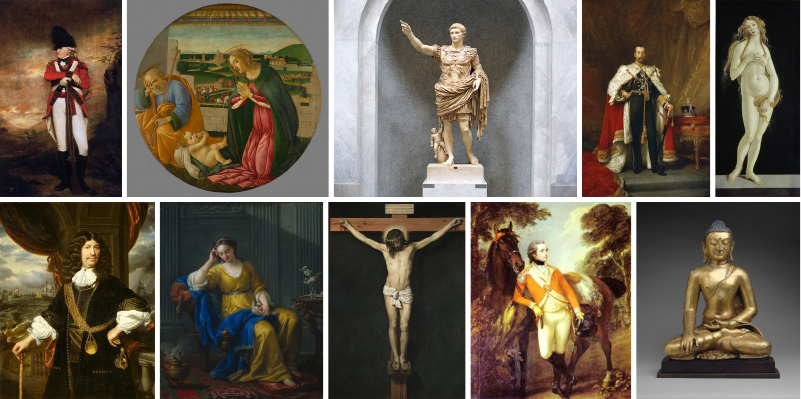}
\caption{Query images for the user study with art-historical poses, e.g., \enquote*{Adlocutio} and \enquote*{Venus pudica.}}
\label{fig:pose-query}
\end{figure}

\section{User Study on Retrieval Results}
\label{chp:study}

In this section, we report the results of a user study that aimed to evaluate the quality of the automatically generated keypoints from a human perspective in a retrieval scenario. We first describe the generation of keypoint descriptors and the experimental setup of the user study before discussing the results.

\textbf{Keypoint descriptors:} For the retrieval task, we convert keypoints into a consistent feature vector representation. 
In doing so, descriptors for the same pose should be nearly identical regardless of position or scale. 
As pose discrimination depends heavily on the relational configuration between body parts \cite{DBLP:journals/tvcg/HoK09}, we do not leverage joint coordinates directly \cite{DBLP:journals/jvca/SoB05, DBLP:conf/humanoids/HaradaTMS04}. 
Instead, we build on \citet{DBLP:journals/tvcg/ChenZNYWX11}
%, who introduced pose descriptors based on the distance, orientation, and angle between individual keypoints and their connections. 
%For our experiments, 
and employ a $52$-dimensional feature descriptor that uses the orientation between two keypoints. 
We obtain $1,515$ images from the ART500k data set not used for training in Section~\ref{chp:exp_boxes}, to which bounding box and keypoint models are applied. 
For each pose, the descriptor from \citet{DBLP:journals/tvcg/ChenZNYWX11} is calculated. 
In addition, we selected $10$ poses with varying art-historical specificity and utilized them as query images (Figure~\ref{fig:pose-query}). 
The small number of examples naturally can only inadequately cover the large variability of relevant body constellations; it is, nonetheless, sufficient to ascertain the models' basic suitability for retrieval tasks.

% For our study, we summarize the responses of $12$ participants who had to rate each search result as \emph{relevant}, \emph{irrelevant}, or \emph{indifferent} compared to the respective query image. 
\textbf{Experimental setup:} For our study, we developed a web interface with detailed instructions for annotation.
A total of $12$ subjects were recruited, personally invited by the participating departments of computer science and art history.
These included seven computer scientists, two art historians, and three persons from other professions. 
% At the beginning of the annotation process, participants had to create a profile and classify themselves in the group of art historians, computer scientists, or other professions.
In the study, several pages were shown, consisting of a query image and %(more or less relevant) 
the corresponding top-20 retrieval results. %examples of the given search result set. 
For each displayed image, participants were asked to vote on whether they thought it was \enquote*{relevant,} \enquote*{irrelevant,} or \enquote*{indifferent} to the query. 
After the questioning, the individual results were ranked in this order: \enquote*{relevant,} \enquote*{indifferent,} and \enquote*{irrelevant.}
%To determine how similar two pose descriptors are, 
We used Euclidean distance to compute a ranking based on the automatically computed descriptors and compared it to the user-generated ranking. 
%we used Euclidean distance with rankings and user-generated scores compared through \ac{NDCG}. 
The results of the user study are reported in Table~\ref{tab:userstudy} and show that our proposed approach also outperforms competing models in retrieval, nevertheless, with decreasing variations between models. 
This can possibly be explained by the fact that it is not necessarily relevant for a user if the alignment of individual keypoints changes as long as the basic pose has very similar meaning. 
However, it may also be that the number of subjects is too small for such conclusions, or that the participants' art-historical knowledge was insufficient to interpret certain details of the poses. 
%For more representative values, the study would have to be repeated with a larger group of users sufficiently proficient in the history of the visual arts. 
In this context, the degree of similarity at which subjects consider poses to be similar is relevant. For instance, one participant excluded crucifixion scenes in which Christ looked to the left rather than downward with his head bowed, as in the query image. % (Figure~\ref{fig:pose-query}, \textit{second row}).

\begin{table}
\caption{Results of the user study on the retrieval of similar poses with \acf{NDCG} as the ranking metric.}
\label{tab:userstudy}

\begin{tabularx}{\linewidth}{@{}XRRRRR@{}}
\toprule
Train set & Stylized & Semi & @5 & @10 & @15 \\
\midrule
COCO 2017 & {\sisetup{round-precision=0}\SI{0}{\percent}} & & \num{0.5626065045880496} & \num{0.5929046028791711} & \num{0.6309289755365093} \\
COCO 2017 & {\sisetup{round-precision=0}\SI{0}{\percent}} & \checkmark & \num{0.5702222757416806}& \num{0.567596068630933} & \num{0.5956858383439985} \\
COCO 2017 & {\sisetup{round-precision=0}\SI{50}{\percent}} & & \num{0.6123574441937504} & \num{0.6053535386709639} & \num{0.6233609099435142} \\
COCO 2017 & {\sisetup{round-precision=0}\SI{50}{\percent}} &\checkmark&\num{0.571346217004936} & \num{0.5900424185793351} & \num{0.6093549589887899} \\
COCO 2017 & {\sisetup{round-precision=0}\SI{100}{\percent}} & & \num{0.5727648495910804} & \num{0.595849882336127} & \num{0.6131130345488688} \\
COCO 2017 & {\sisetup{round-precision=0}\SI{100}{\percent}} & \checkmark & \num{0.5844929687342831}& \num{0.606948907593188} & \num{0.6303804494689661} \\
\midrule
PoPArt & {\sisetup{round-precision=0}\SI{0}{\percent}} & & \num{0.5674734449562145} & \num{0.5722388642863352} & \num{0.5942983606417748} \\
PoPArt & {\sisetup{round-precision=0}\SI{0}{\percent}} & \checkmark & \textbf{\num{0.6413402576973029}} & \textbf{\num{0.6204677429158516}} & \textbf{\num{0.6343858477271396}} \\
\bottomrule
\end{tabularx}
\end{table}

\section{Conclusions and Future Work}
\label{chp:conc}

In this paper, we have investigated domain adaption techniques to estimate human poses in art-historical images. 
Therefore, we have suggested a two-stage approach based on two Transformer models that utilizes a semi-supervised teacher-student design. 
To reduce the gap between photographs of real-world objects and the art domain, we augmented images depicting real-world scenes with unlabeled, domain-specific data. 
Moreover, we introduced a reasonably large art-historical data set called \acf{PoPArt} to systematically test the validity of human pose estimators. 
Comparisons with more common approaches that use pre-trained models or adapt existing data sets with style transfer indicated that performance can be further improved with unlabeled data. 
While it is not necessary to annotate large amounts of art-historical material, it is essential to include at least smaller, domain-specific labeled data in the training procedure, rather than relying solely on synthetically generated imagery. 
Depending on the test set, models trained entirely or partially with style transfer underperform in \acl{AP} by between $7.32$ to $28.12$\,\% for person detection and between $27.33$ to $28.15$\,\% for keypoint prediction, even with semi-supervised learning. 
A user study has confirmed the feasibility of the proposed approach for retrieval tasks, thus also enabling the search for resembling poses of human figures; however, in this case the difference with other models performance-wise is smaller.

% Since we wanted to explicitly demonstrate how semi-supervised learning can improve the performance of human pose estimation in the visual arts, we did not use the best performing network architectures in these experiments. 
In the future, we intend to analyze the potential of recently introduced Transformer models, such as the Pyramid Vision Transformer %\ac{PVT} 
presented by \citet{DBLP:journals/corr/abs-2106-13797}. Further improvement of the training process could be achieved by applying style transfer to unlabeled instead of only labeled data. We also plan to extend the \ac{PoPArt} data set with additional bounding boxes, enhancing its usefulness for training person detection models.

\section*{Acknowledgements}

We thank the participants in the retrieval study for their valuable contributions.

\clearpage
\bibliographystyle{ACM-Reference-Format}
\balance
\bibliography{main}
\balance

\clearpage
\onecolumn
\setcounter{section}{0}

\renewcommand\thesection{\Alph{section}}

\section{Pose estimation with ground truth bounding boxes}

\begin{table*}[hp]
\caption{Keypoint detection results on the PoPArt test set with annotated ground-truth bounding boxes. $AP_{S}$ is neglected as no test data is available for small human figures. The best performing approach is bold.}
\label{tab:exp_keypoints}
\begin{tabularx}{\textwidth}{@{}XXRRRRRRRR@{}}
\toprule
Test set & Train set & Stylized & Semi & $AP$ & $AP_{50}$ & $AP_{75}$ & $AP_{M}$ &$AP_{L}$ & $AR$  \\
\midrule
PoPArt & COCO 2017 & {\sisetup{round-precision=0}\SI{0}{\percent}} & & \num{0.6969} & \num{0.9024} & \num{0.7869} & \num{0.4612} & \num{0.702} & \num{0.7614} \\
& COCO 2017 & {\sisetup{round-precision=0}\SI{0}{\percent}} & \checkmark &\num{0.6724} & \num{0.8961} & \num{0.7343} & \num{0.4946} & \num{0.6764} & \num{0.7911} \\
& COCO 2017 & {\sisetup{round-precision=0}\SI{50}{\percent}} & & \num{0.7058} & \num{0.9187} & \num{0.7691} & \num{0.494} & \num{0.7113} & \num{0.7868} \\
& COCO 2017 & {\sisetup{round-precision=0}\SI{50}{\percent}} & \checkmark & \num{0.6943} & \num{0.9203} & \num{0.7594} & \num{0.4795} & \num{0.7031} & \num{0.8031} \\
& COCO 2017 & {\sisetup{round-precision=0}\SI{100}{\percent}} & & \num{0.6981} & \num{0.9178} & \num{0.7776} & \num{0.54} & \num{0.7047} & \num{0.783} \\
& COCO 2017 & {\sisetup{round-precision=0}\SI{100}{\percent}} & \checkmark & \num{0.7005} & \num{0.9254} & \num{0.7826} & \num{0.5449} & \num{0.7073} & \num{0.8016} \\
\cmidrule{2-10}
& PoPArt & {\sisetup{round-precision=0}\SI{0}{\percent}} & & \num{0.7552} & \textbf{\num{0.9474}} & \textbf{\num{0.8488}} & \textbf{\num{0.5941}} & \num{0.7603} & \num{0.8297} \\
& PoPArt & {\sisetup{round-precision=0}\SI{0}{\percent}} & \checkmark & \textbf{\num{0.7592}} & \num{0.9362} & \num{0.8395} & \num{0.4884} & \textbf{\num{0.7676}} & \textbf{\num{0.8459}}\\
\bottomrule
\end{tabularx}
\end{table*}

\noindent To evaluate only the keypoint detection stage, we assess the models' performance based on the ground-truth bounding boxes from the \ac{PoPArt} test set. As the results are not affected by bounding box assignment, we can thus estimate the best possible prediction outcome. Table~\ref{tab:exp_keypoints} shows the results of each method. We observe that the semi-supervised learning technique does not affect \ac{AP}, or only to a small extent; \ac{AR}, on the other hand, increases for all methods. Moreover, we notice that keypoint prediction performs significantly better with ground truth boxes than with those from an upstream bounding box detection; \ac{AP} of the best model increases from \num{0.5258} to \num{0.7592} and \ac{AR} from \num{0.7464} to \num{0.8459}. This shows that especially person localization has to be improved to achieve better results.  
\vspace{12pt}

\section{Additional predictions from pose estimators}

\captionsetup[subfigure]{labelformat=empty}
\begin{figure*}[hp]
\begin{center}

\centering
\begin{subfigure}{.09\linewidth}
  \centering
  \includegraphics[width=\linewidth]{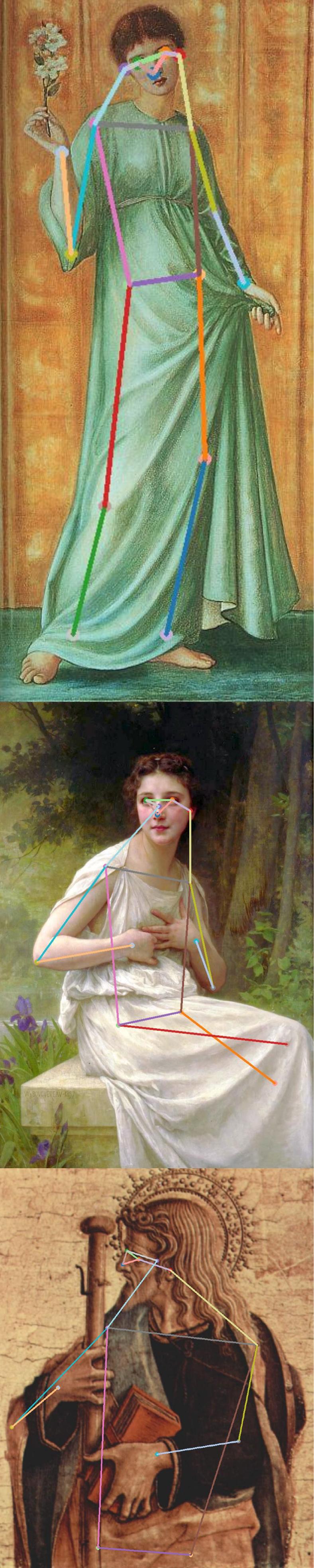}
  \caption{(a)}
  \label{fig:examples-gt}
\end{subfigure}%
\hfill
\begin{subfigure}{.09\linewidth}
  \centering
  \includegraphics[width=\linewidth]{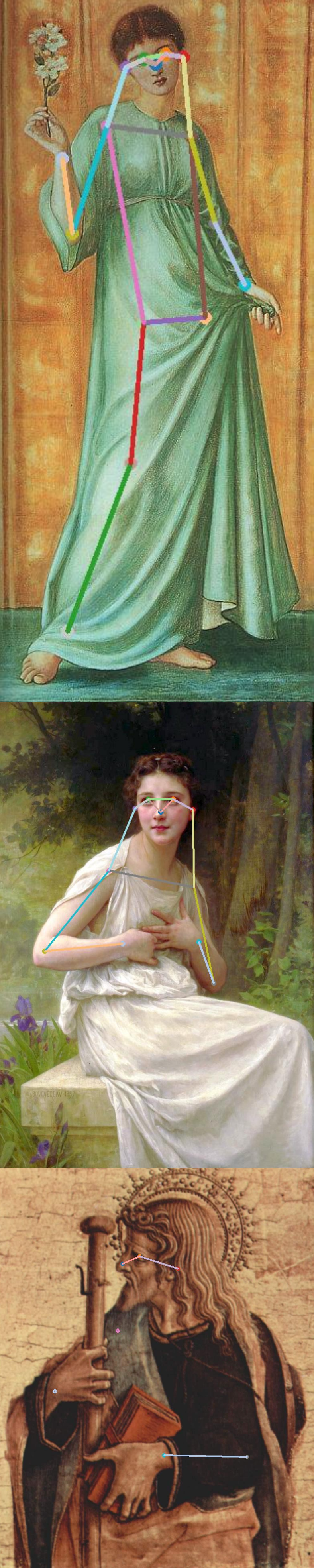}
  \caption{(b)}
  \label{fig:examples-openpose}
\end{subfigure}%
\hfill
\begin{subfigure}{.09\linewidth}
  \centering
  \includegraphics[width=\linewidth]{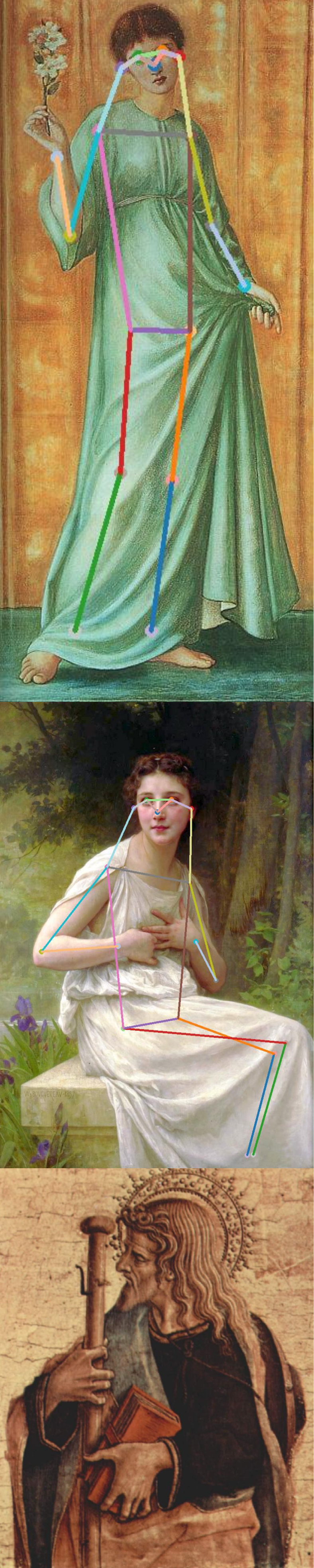}
  \caption{(c)}
  \label{fig:examples-coco}
\end{subfigure}%
\hfill
\begin{subfigure}{.09\linewidth}
  \centering
  \includegraphics[width=\linewidth]{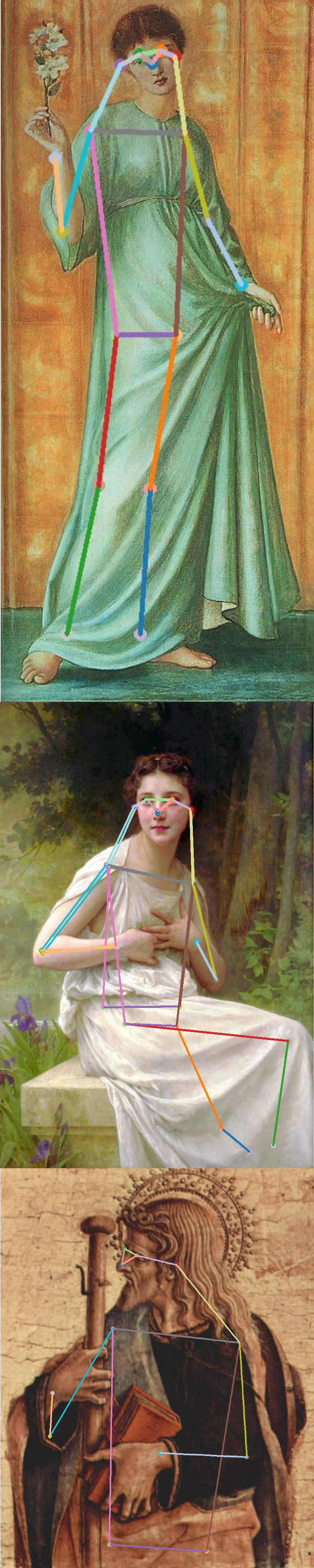}
  \caption{(d)}
  \label{fig:examples-our}
\end{subfigure}%
\hfill
\begin{subfigure}{.09\linewidth}
  \centering
  \includegraphics[width=\linewidth]{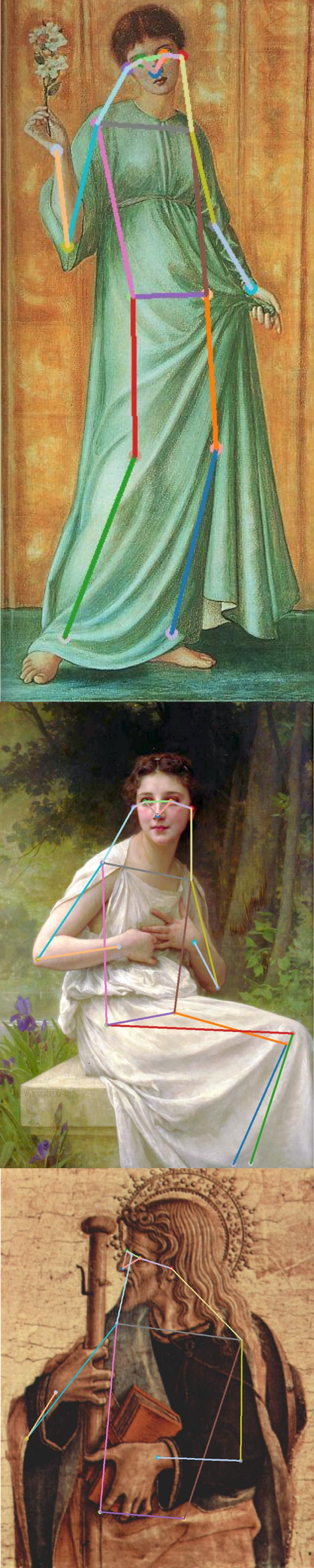}
  \caption{(e)}
  \label{fig:examples-our}
\end{subfigure}%
\hspace{17.5pt}
\hfill
\hspace{17.5pt}
\begin{subfigure}{.09\linewidth}
  \centering
  \includegraphics[width=\linewidth]{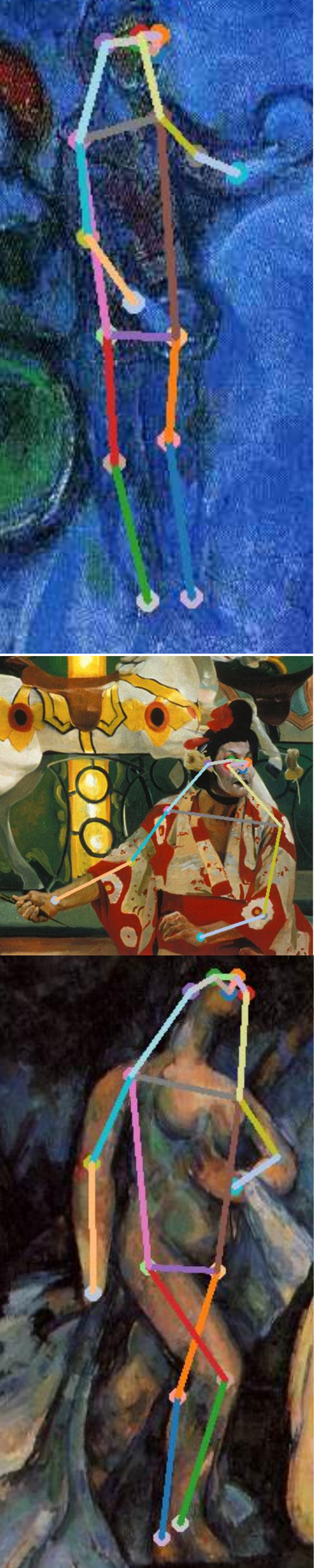}
  \caption{(a)}
  \label{fig:examples-gt}
\end{subfigure}%
\hfill
\begin{subfigure}{.09\linewidth}
  \centering
  \includegraphics[width=\linewidth]{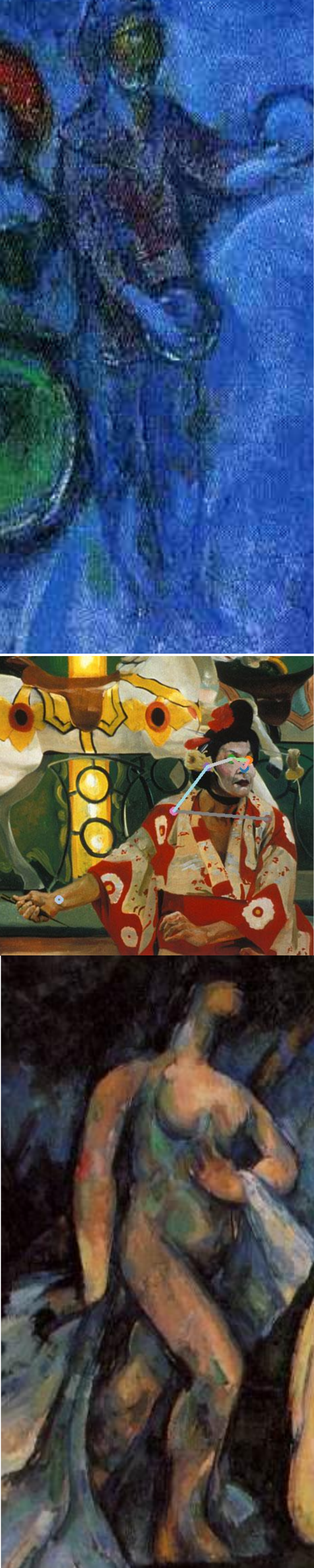}
  \caption{(b)}
  \label{fig:examples-openpose}
\end{subfigure}%
\hfill
\begin{subfigure}{.09\linewidth}
  \centering
  \includegraphics[width=\linewidth]{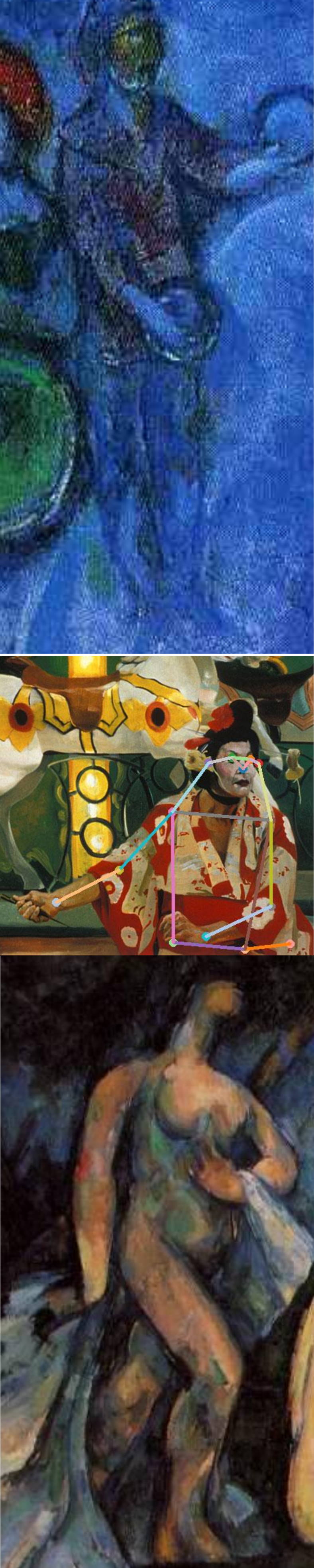}
  \caption{(c)}
  \label{fig:examples-coco}
\end{subfigure}%
\hfill
\begin{subfigure}{.09\linewidth}
  \centering
  \includegraphics[width=\linewidth]{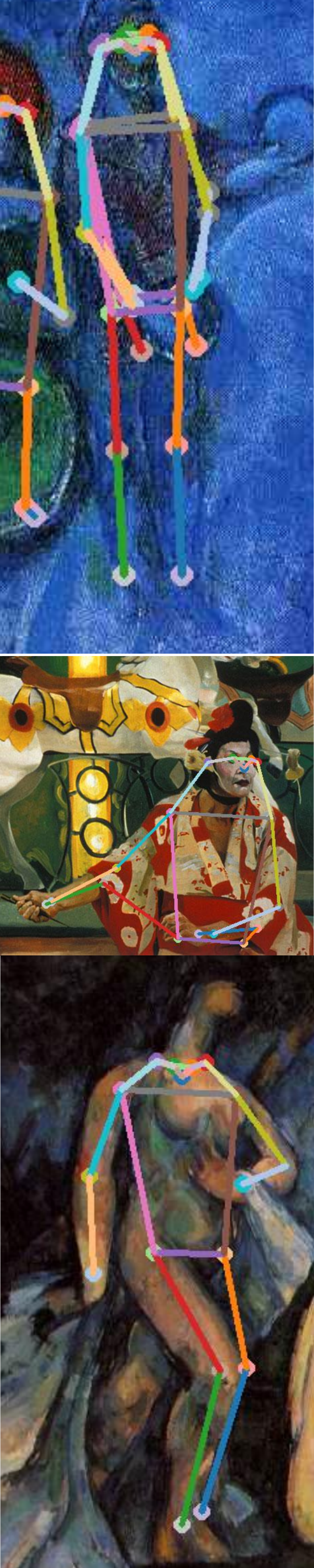}
  \caption{(d)}
  \label{fig:examples-our}
\end{subfigure}%
\hfill
\begin{subfigure}{.09\linewidth}
  \centering
  \includegraphics[width=\linewidth]{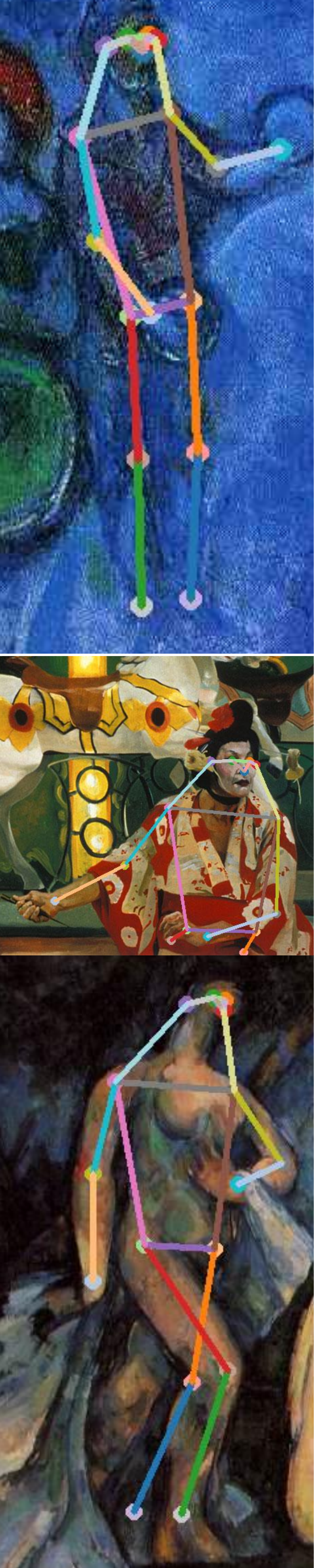}
  \caption{(e)}
  \label{fig:examples-our}
\end{subfigure}%
\end{center}

\caption{Predictions of the different models are overlaid on examples from the \ac{PoPArt} test data: (a) ground-truth annotations; (b) estimations of OpenPose; (c) predictions of the model trained with COCO 2017 without style transfer and without semi-supervised training; (d) predictions with style transfer; (e) \ac{PoPArt} trained with semi-supervised learning.}
\label{fig:two_stages}
\end{figure*}

\end{document}